\documentclass[twocolumn]{IEEEtran}
\usepackage{lscape}
\usepackage{booktabs}
\usepackage{authblk}
\usepackage{url}
\usepackage{silence}
\usepackage[font=scriptsize]{caption}

\usepackage[pdftex]{graphicx}
\usepackage{epstopdf}
\usepackage{pifont}
\usepackage[table,xcdraw]{xcolor}
\usepackage{array}
\usepackage{verse}
\usepackage{longtable}
\usepackage{supertabular}
\usepackage{cite}
\usepackage{color}
\usepackage{amsmath}
\usepackage{lettrine}
\usepackage{hyperref}
\usepackage{multirow}
\usepackage{multicol}
\usepackage{dirtytalk}
\usepackage{color}

\begin{document}
\title{Secure and Robust Machine Learning for Healthcare Applications: A Survey}
\title{Secure and Robust Machine Learning for Healthcare: A Survey}

\author{Adnan Qayyum$^1$\thanks{Email: adnan.qayyum@itu.edu.pk}, Junaid Qadir$^1$, Muhammad Bilal$^2$, and Ala Al-Fuqaha$^3$ \\ \vspace{2mm}
$^1$ Information Technology University (ITU), Punjab, Lahore, Pakistan \\ 
$^2$ University of the West England (UWE), Bristol, United Kingdom \\
$^3$ Hamad Bin Khalifa University (HBKU), Doha, Qatar}

\maketitle

\begin{abstract}
\label{abstract}
Recent years have witnessed widespread adoption of machine learning (ML)/deep learning (DL) techniques due to their superior performance for a variety of healthcare applications ranging from the prediction of cardiac arrest from one-dimensional heart signals to computer-aided diagnosis (CADx) using multi-dimensional medical images. Notwithstanding the impressive performance of ML/DL, there are still lingering doubts regarding the robustness of ML/DL in healthcare settings (which is traditionally considered quite challenging due to the myriad security and privacy issues involved), especially in light of recent results that have shown that ML/DL are vulnerable to adversarial attacks. In this paper, we present an overview of various application areas in healthcare that leverage such techniques from security and privacy point of view and present associated challenges. In addition, we present potential methods to ensure secure and privacy-preserving ML for healthcare applications. Finally, we provide insight into the current research challenges and promising directions for future research.  
\end{abstract}

\section{Introduction}
We are living in the age of algorithms, in which machine learning (ML)/deep learning (DL) systems have transformed multiple industries such as manufacturing, transportation, and governance. Over the past few years, DL has provided state of the art performance in different domains---e.g., computer vision, text analytics, and speech processing, etc. Due to the extensive deployment of ML/DL algorithms in various domains (e.g., social media), such technology has become inseparable from our routine life. ML/DL algorithms are now beginning to influence healthcare as well---a field that has traditionally been impervious to large-scale technological disruptions \cite{latif20175g}. ML/DL techniques have shown outstanding results recently in versatile tasks such as recognition of body organs from medical images \cite{yan2016multi}, classification of interstitial lung diseases \cite{anthimopoulos2016lung}, detection of lungs nodules \cite{shen2015multi}, medical image reconstruction \cite{schlemper2017deep,mehta2017rodeo}, and brain tumor segmentation \cite{havaei2017brain}, to name a few.



It is highly expected that intelligent software will assist radiologists and physicians in examining patients in the near future \cite{bourzac2013computer} and ML will revolutionize the medical research and practice \cite{xing2018artificial}. Clinical medicine has emerged as a exciting application area for ML/DL models, and these models have already achieved human-level performance in clinical pathology \cite{bejnordi2017diagnostic}, radiology \cite{rajpurkar2017chexnet}, ophthalmology \cite{gulshan2016development}, and dermatology \cite{esteva2017dermatologist}. Some of these studies have even reported that DL models outperform human physicians on average. The aspect of better performance of DL models in comparison with humans has led to the development of computer-aided diagnosis systems---for instance, U.S. Food and Drug Administration has announced the approval of an intelligent diagnosis system for medical images that will not require any human intervention\footnote{\url{https://tinyurl.com/FDA-AI-diabetic-eye}}. 

The potential of ML models for healthcare applications is also benefitting from the progress in concomitantly-advancing technologies like cloud/edge computing, mobile communication, and big data technology \cite{latif2018automating}. Together with these technologies, ML/DL is capable of producing highly accurate predictive outcomes and can facilitate the human-centered intelligent solutions \cite{chen2014big}. Along with other benefits like enabling remote healthcare services for rural and low-income zones, these technologies can play a vital role in revitalizing the healthcare industry. 


Notwithstanding the impressive performance of DL algorithms, many recent studies have raised concerns about the security and robustness of ML models---for instance, Szegedy et al. demonstrated for the first time that DL models are strictly vulnerable to carefully crafted adversarial examples \cite{szegedy2013intriguing}. Similarly, various types of data and model poisoning attacks have been proposed against DL systems \cite{shafahi2018poison} and different defenses against such strategies have been proposed in the literature \cite{yuan2019adversarial}. However, the robustness of defense methods is also questionable and different studies have shown that most of the defense techniques fail against a particular attack. The discovery of the fact that DL models are neither secure nor robust hinders significantly their practical deployment in security-critical applications like predictive healthcare which is essentially life-critical. For instance, researchers have already demonstrated the threat of adversarial attacks on ML-based medical systems \cite{finlayson2019adversarial,papangelou2018toward}. Therefore, ensuring the integrity and security of DL models and health data are paramount to the widespread adoption of ML/DL in the industry. 

\begin{figure*}[!ht]
    \centering
    \includegraphics[width=0.85\textwidth]{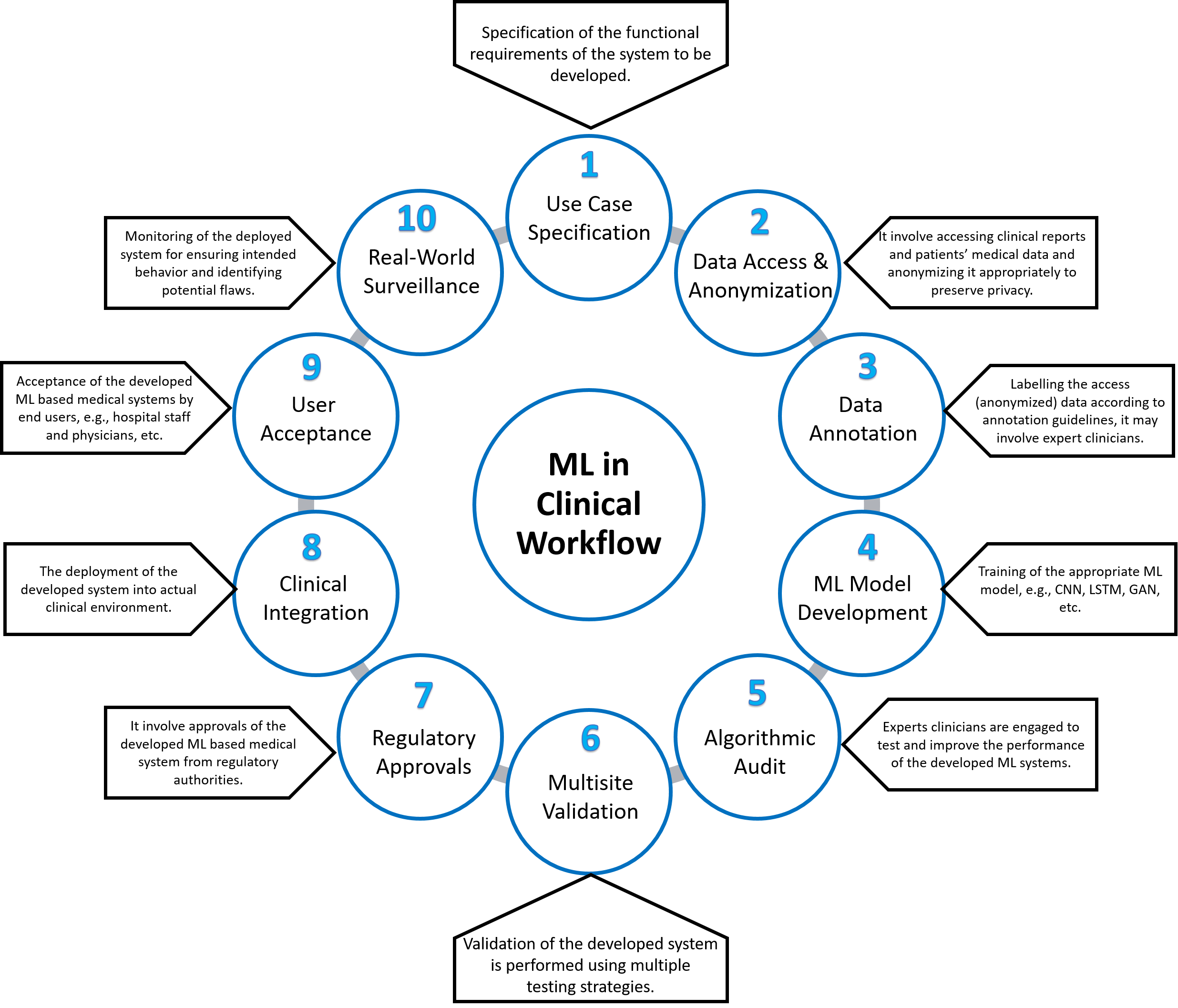}
    \caption{The illustration of major phases for development of machine learning (ML) based healthcare systems.}
    \label{fig:ml_workflow}
\end{figure*}

In this paper, we present a comprehensive survey of existing literature on the security and robustness of ML/DL models with a specific focus on their applications in healthcare systems. We also highlight various challenges and sources of vulnerabilities that hinder the robust application of ML/DL models in healthcare applications. In addition, potential solutions to address these challenges are presented in this paper. In summary, the following are the specific contributions of this paper.


\begin{enumerate}
    \item We present an overview of different applications of ML/DL models in healthcare. 
    \item We formulate the ML pipeline for predictive healthcare and identify various sources of vulnerabilities at each stage.
    \item We highlight various conventional security and privacy-related challenges as well as ones that arise with the adoption of ML/DL models. 
    \item We present potential solutions for the robust application of ML/DL techniques for healthcare applications. 
    \item Finally, we highlight various open research issues that require further investigation. 
\end{enumerate}

\textit{Organization of the Paper:} The rest of the paper is organized as follows. In Section \ref{sec:ml_apps}, various applications of ML and DL techniques in healthcare are discussed. Section \ref{sec:challenges} presents the ML pipeline in data-driven healthcare and various sources of vulnerabilities along with different challenges associated with the use of ML. Different potential solutions to ensure secure and privacy-preserving ML are discussed in Section \ref{sec:solutions} and various open research issues are outlined in Section \ref{sec:open}. Finally, we conclude the paper in Section \ref{sec:con}.    


\section{ML for Healthcare: Applications}
\label{sec:ml_apps}
In this section, various prominent applications of ML in healthcare are discussed and we start by providing the big picture of ML in the context of healthcare.

\subsection{ML in Healthcare: The Big Picture}
The major phases for developing a ML-based healthcare system are illustrated in Figure \ref{fig:ml_workflow} and major types of ML/DL that can be used in healthcare applications are briefly described next. 

\subsubsection{Unsupervised Learning} The ML techniques utilizing unlabelled data are known as unsupervised learning methods. Widely used examples of unsupervised learning methods are a clustering of data points using a similarity metric and dimensionality reduction to project high dimensional data to lower-dimensional subspaces (sometimes also referred to as feature selection). In addition, unsupervised learning can be used for anomaly detection, e.g., clustering \cite{chandola2009anomaly}. Classical examples of unsupervised learning methods in healthcare include the prediction of heart diseases using clustering \cite{pandey2013datamining} and prediction of hepatitis disease using principal component analysis (PCA) which is a dimensionality reduction technique \cite{polat2007prediction}.

\subsubsection{Supervised Learning} Such methods that build or map the association between the inputs and outputs using labeled training data are characterized as supervised learning methods. If the output is discrete then the task is called \textit{classification} and for a continuous value output, the task is called \textit{regression}. Classical examples of supervised learning methods in healthcare include the classification of different types of lung diseases (nodules) \cite{shen2015multi} and recognition of different body organs from medical images \cite{yan2016multi}. Sometimes, ML methods can be neither supervised nor unsupervised, i.e., where the training data contains both labeled and unlabelled samples. Methods utilizing such data are known as semi-supervised learning methods. A systematic review of supervised and unsupervised learning techniques can be found in \cite{alloghani2020systematic}.

\subsubsection{Semi-supervised Learning} Semi-supervised learning methods are useful when both labelled and unlabelled samples are available for training, typically, a small amount of labelled data and a large amount of unlabelled data. Semi-supervised learning techniques can be particularly useful for a variety of healthcare applications as acquiring a sufficient amount of labelled data for model training is difficult in healthcare. Different facets of semi-supervised learning using different learning techniques have been proposed in the literature. For instance, a semi-supervised clustering method for healthcare data is presented in \cite{sohail2019euclidean} and a semi-supervised ML approach for activity recognition using sensors data is presented in \cite{zahin2019sensor}. In \cite{mahapatra2017semi,bai2017semi}, authors applied a semi-supervised learning method to medical image segmentation.  

\subsubsection{Reinforcement Learning} Methods that learn a policy function given a set of observations, actions, and rewards in response to actions performed over time fall in the class of reinforcement learning (RL) \cite{sutton1998introduction}. RL has a great potential to transform many healthcare applications and recently, it has been used for context-aware symptoms checking for disease diagnosis \cite{kao2018context}. Furthermore, the potential of using RL for healthcare applications can be seen through the recent example of the Go game, where a computer using RL with the integration of supervised and unsupervised learning methods defeated a human champion player \cite{silver2016mastering}. 


\subsection{Applications of ML in Healthcare}
\begin{figure}[!h]
    \centering
    \includegraphics[width=0.45\textwidth]{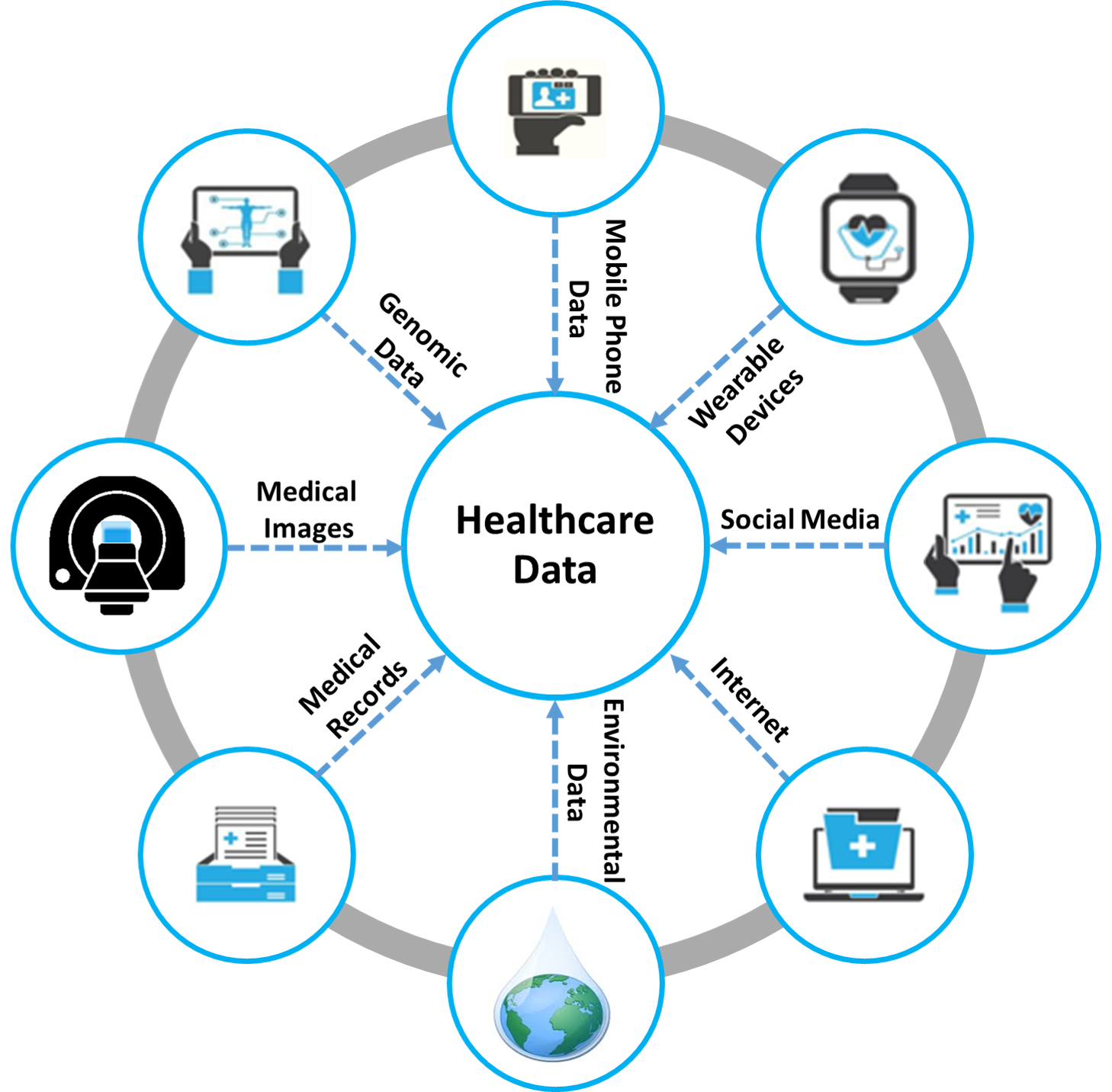}
    \caption{Illustration of heterogeneous sources contributing to healthcare data.} 
    \label{fig:sources}
\end{figure}

Healthcare service providers generate a large amount of heterogeneous data and information daily, making it difficult for the ``traditional methods'' to analyze and process it. ML/DL methods help to effectively analyze this data for actionable insights. In addition, there are heterogeneous sources of data that can augment healthcare data such as genomics, medical data, data from social media, and environmental data, etc. A depiction of these sources of data is shown in Figure \ref{fig:sources}. The four major applications of healthcare that can benefit from ML/DL techniques are prognosis, diagnosis, treatment, and clinical workflow, which are described next. 

\subsubsection{Applications of ML in Prognosis}
Prognosis is the process of predicting the expected development of a disease in clinical practice. It also includes identification of symptoms and signs related to a specific disease and whether they will become worse, improve, or remain stable over time and identification of potential associated health problems, complications, ability to perform routine activities, and the likelihood of survival. As in clinical setting, multi-modal patients' data is collected, e.g., phenotypic, genomic, proteomic, pathology tests results, and medical images, etc., which can empower the ML models to facilitate disease prognosis, diagnosis and treatment \cite{collins2018machine}. For instance, ML models have been largely developed for the identification and classification of different types of cancers, e.g., brain tumor \cite{afshar2018brain} and lung nodules \cite{zhu2018deeplung}. However, the potential applications ML for disease prognosis, i.e., prediction of disease symptoms, risks, survivability, and recurrence have been exploited under recent translational research efforts that aim to enable personalized medicine. However, the field of personalized medicine is nascent that requires extensive development of adjacent fields like bioinformatics, strong validation strategies, and demonstrably robust applications of ML thus to achieve the huge and translational impact.   
 
\subsubsection{Applications of ML in Diagnosis}

\paragraph{Electronic Health Records (EHRs)}
Hospitals and other healthcare service providers are producing a large collection of electronic health records (EHRs) on a daily basis and comprise of structured and unstructured data that contains a complete medication history of patients. ML-based methods have been utilized for the extraction of clinical features for facilitating the diagnosis process \cite{jensen2012mining}. For example, a semi-supervised approach for the extraction of diagnosis information from unstructured EHRs is presented in \cite{wang2012extracting}. The use of ML for the diagnosis of diabetes from EHRs is presented in \cite{zheng2017machine}. In \cite{nestor2019feature}, features robustness using EHRs data for the year of care for each record is examined for two tasks, i.e., mortality prediction and length-of-stay and authors showed that prediction performance gets degraded when ML models are trained on historical data and tested on unseen (future) data.

\begin{figure*}
    \centering
    \includegraphics[width=0.9\textwidth]{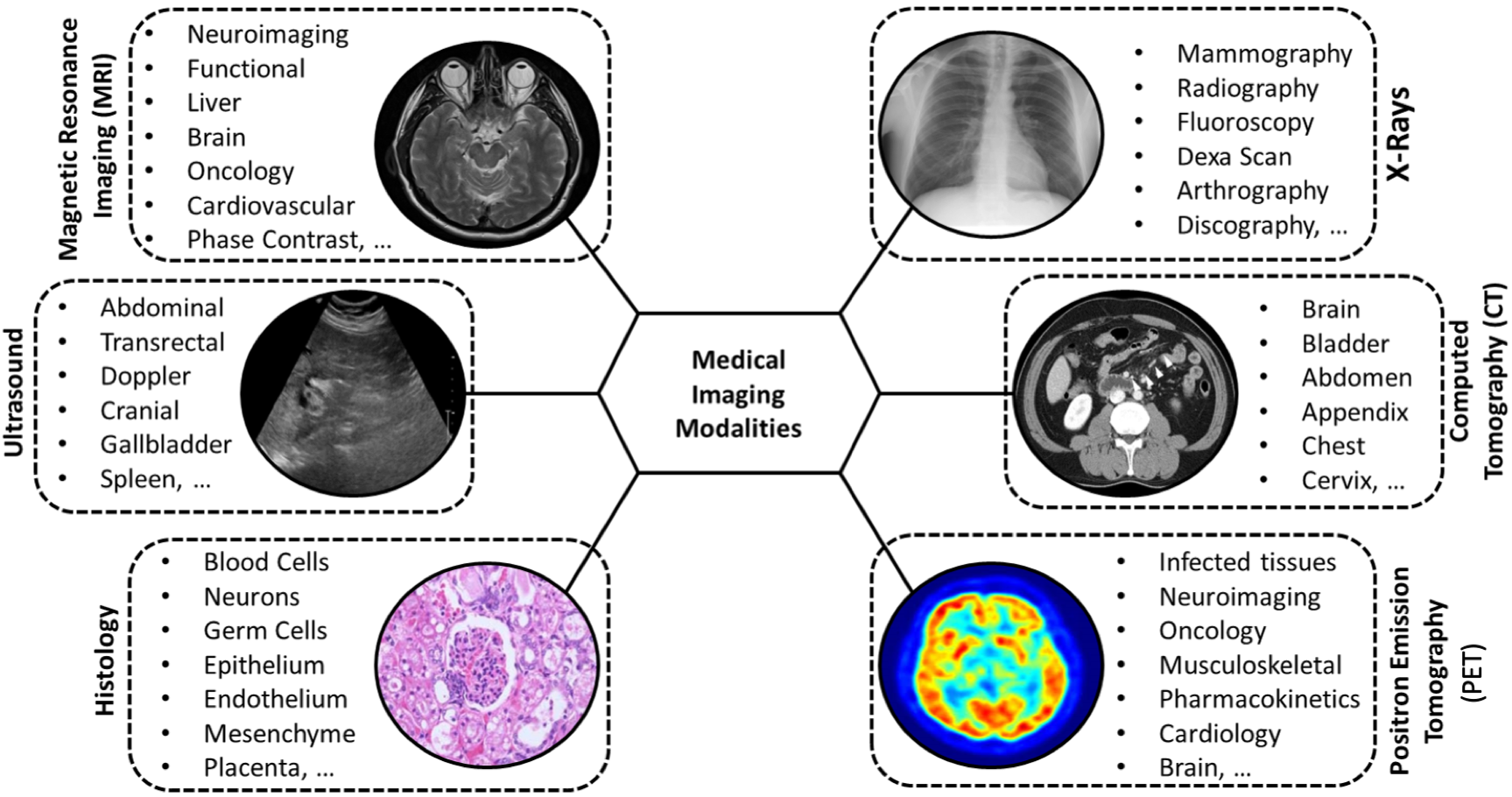}
    \caption{A typology of commonly used medical imaging modalities (adapted from \cite{anwar2018medical}).}
    \label{fig:modalities}
\end{figure*}

\paragraph{ML in Medical Image Analysis} 
In medical image analysis, ML techniques are used for efficient and effective extraction of information from medical images that are acquired using different imaging modalities such as magnetic resonance imaging (MRI), computed tomography (CT), ultrasound, and positron emission tomography (PET), etc. These modalities provide important functional and anatomical information about different body organs and play a crucial role in the detection/localization and diagnosis of abnormalities. A taxonomy of key medical imaging modalities is presented in Figure \ref{fig:modalities}. The key purpose of medical image analysis is to assist clinicians and radiologists for efficient diagnosis and prognosis of the diseases. The prominent tasks in medical image analysis include detection, classification, segmentation, retrieval, reconstruction, and image registration which are discussed next. Moreover, fully automated intelligent medical image diagnosis systems are expected to be part of next-generation healthcare systems.  



\begin{itemize}
    \item \textit{Enhancement:} Enhancement of degraded medical images is an important pre-processing step that directly effects the diagnosis process. There are many sources of noise and disturbances encountered in the medical image acquisition process which degrade the quality and significance of the resultant images. For instance, generating MRI images is a quite lengthy process that typically requires several minutes to produce a good quality image and to acquire detailed soft-tissue contrast, patients have to remain still and straight as much as possible. Because movements can cause false artifacts in image acquisition, the complete process has to be repeated usually multiple times to produce significantly useful images. Also, depending on the body area being scanned and the number of images to be taken, patients might be asked to hold their breath during short scans \cite{lustig2008compressed}. Therefore, any movement of the subject can introduce artifacts in the acquired image. Moreover, some sort of mechanical noise is also sometimes introduced in the output image. In the literature, different DL models are used for denoising medical images such as convolutional denoising autoencoders \cite{gondara2016medical} and GANs. In addition, GANs have been successfully used for cleaning motion artifacts introduced in multi-shot MRI images \cite{latif2018automating}. Super-resolution is yet another powerful and impactful enhancement technique for medical images, e.g., MRI denoising \cite{chen2018brain}. 
    \item \textit{Detection:} The process of identifying specific disease patterns or abnormalities (e.g., tumor, cancer) in medical images is known as detection. In traditional clinical practice, such abnormalities are identified by expert radiologists or physicians that often require a lot of time and effort. Whereas, DL based methods have shown their potential for this task and various studies have been presented in the literature for the detection of diseases. For instance, a locality-sensitive approach utilizing CNN for the detection and classification of nuclei colon cancer in histopathological images is presented in \cite{sirinukunwattana2016locality}. A hybrid method utilizing handcrafted features and a CNN model for the detection of mitosis in breast cancer images is presented in \cite{wang2014mitosis}.
    \item \textit{Classification} DL models in particular, convolutional neural networks (CNNs) have proven to give high performance in medical image classification tasks when compared with other state-of-the-art non-learning based techniques. Modality classification, recognizing different body organs, and abnormalities from medical images using CNNs have been extensively studied in the literature. In \cite{yan2016multi}, an approach using CNN for multi-instance recognition of different body organs is presented and a CNN based method for classification of interstitial lung diseases (ILDs) is presented in \cite{anthimopoulos2016lung}. In another study, a CNN model is trained for the classification of lung nodules \cite{shen2015multi}. 
    
    Transfer learning approaches have also been used for medical image classification \cite{yu2017deep}. In transfer learning, a pre-trained DL model (typically trained on natural images) is fine-tuned on a comparatively small dataset of medical images. The results obtained by this approach, as reported in the literature, are promising; however, a few studies have reported contradictory results. For instance, results obtained by transfer learning in \cite{antony2016quantifying} and \cite{kim2016deep} are contradictory. 
    
    \item \textit{Segmentation:} The segmentation of tissues and organs in medical images enables quantitative analysis of abnormalities in terms of clinical parameters, e.g., automatically measuring the volume and shape of cancer in brain images. In addition, the extraction of such clinically significant features is an important and foremost step in computer-aided detection and diagnosis systems that we discuss later in this section. The process of segmentation deals with the partitioning of an image into multiple non-overlapping parts using a pre-defined criterion such as intrinsic color, texture, and contrast, etc. Addressing the problem of segmentation utilizing various DL models (e.g, CNN and recurrent neural network (RNN) \cite{stollenga2015parallel}) is widely studied in the literature and the common architecture used for segmentation of medical images is U-net \cite{ronneberger2015u}. Various DL architectures are being proposed for the segmentation of multi-modal images such as the brain, skin cancer, CT images, etc. as well as segmentation of volumetric images \cite{milletari2016v}. An overview of various DL models for segmentation of medical images is presented in \cite{hesamian2019deep}. 
    
    \item \textit{Reconstruction:} The process of generating interpretable images from raw data acquired from the imaging sensor is known as medical image reconstruction. The fundamental problem in medical image reconstruction is to accelerate the inherently slow data acquisition process, which is an interesting ill-posed inverse problem in which we want to determine the system's input given its output. Many important medical imaging modalities require a lot of time for reconstructing an image from the raw data samples, e.g., MRI and CT. Thus in medical image reconstruction, we aim to reduce image acquisition time and storage space.
    
    Research on medical image reconstruction using deep models is drastically increasing and various DL models such as CNNs \cite{chen2017low} and autoencoders \cite{mehta2017rodeo} have been extensively used for the reconstruction of MRI and CT images. Recently, generative adversarial networks (GANs) have been widely used for the reconstruction of medical images and have produced outstanding results. For instance, a GAN based MRI reconstruction method is presented in \cite{usman2019motion} that also cleans the motion artifacts. 
    
    \item \textit{Image Registration:} Image registration is the process of mapping input images with respect to a reference image and it is the first step in image fusion. Image registration has many potential applications in medical image analysis as described in detail by El-Gamal et al. \cite{el2016current}, however, their use in actual clinical applications is very limited \cite{ker2017deep}. To facilitate the surgical spinal screw implant or tumor removal, image registration is usually applied in spinal surgery or neurosurgery for the localization of spinal bony landmark or a tumor, respectively. Various similarity metrics and reference points are calculated to align the sensed image with the reference image. In \cite{yang2017quicksilver}, a framework for deformable image registration named as Quicksilver is proposed that uses the large deformation diffeomorphic metric mapping (LDDMM) model for patch-wise prediction strategy. Similarly, an unsupervised learning based methods for deformable image registration is presented in . In \cite{miao2016cnn}, a CNN based regression approach for 2D/3D image registration is presented that addresses two fundamental limitations of existing intensity-based image registration methods, i.e., small capture range and slow computation.

    \item \textit{Retrieval:} The recent era has witnessed the revolution of digital interventions from the large-scale image and video collections to big data. This trend is true for medical imaging as well, every hospital and clinic having radiology services are producing thousands of medical images daily in diverse modalities, resulting in the growth of large-scale multi-modal medical image repositories. Thus making it difficult to manage and query such huge databases. In particular, it is more challenging for multi-modal medical data. To facilitate the production and management of multi-modal medical data, traditional methods are not sufficient and various ML/DL techniques are proposed in the literature \cite{shen2017deep,qayyum2017medical}. 
    
    In routine practice, clinicians usually compare the current cases with the previous ones, mainly to effectively plan the diagnosis and treatment of the patient being examined. In this regard, identifying modality (i.e., modality classification discussed above) is of great significance as it serves as an initial tool to facilitate the process of comparison and an efficient modality classification system will reduce the search space by only looking for relevant images in the collections of the desired modality. 
\end{itemize}



\subsubsection{Applications of ML in Treatment}
\paragraph{Image Interpretation}
As discussed above, medical images are widely used in the routine clinical practice and the analysis and interpretation of these images are performed by expert physicians and radiologists. To narrate the findings regarding images being studied, they write textual radiology reports about each body organ that was examined in the conducted study. However, writing such reports is very challenging in some scenarios, e.g., less experienced radiologists and healthcare service providers in rural areas where the quality of healthcare services is not up to the mark. On the other side, for experienced radiologists and pathologists, the process of preparing high-quality reports can be tedious and time-consuming which can be exacerbated by a large number of patients visiting daily. Therefore, various researchers have attempted to address this problem using natural language processing (NLP) and ML techniques. In \cite{zech2018natural}, a natural language processing based method is proposed for annotating clinical radiology reports. A multi-task ML based framework is proposed for automatic tagging and description of medical images \cite{jing2017automatic}. In a similar study \cite{wang2018tienet}, an end-to-end architecture developed with the integration of CNN and RNN is presented for thorax disease classification and reporting in chest X-rays. In \cite{xue2018multimodal}, a novel multi-modal model utilizing CNN and long short term memory (LSTM) network is developed for automatic report generation. 

\paragraph{ML in Real-time Health Monitoring}
Real-time monitoring of critical patients is crucial and is a key component of the treatment process. Continuous health monitoring using wearable devices, IoT sensors, and smartphones is gaining interest among people. In a typical setting of continuous health monitoring, health data is collected using a wearable device and smartphone and then transmitted to the cloud for analysis using an ML/DL technique. Then the outcomes are transmitted back to the device for appropriate action(s). For instance, a framework having a similar system architecture is presented in \cite{jindal2016integrating}. The system is developed by integrating mobile and cloud for monitoring of heart rate using PPG signals. Similarly, a review of different ML techniques for human activity recognition with application to remote monitoring of patients using wearable devices is presented in \cite{attal2015physical}. The sharing of health data with clouds for further analysis raises many privacy and security challenges that we discuss in the next section.

\subsubsection{Applications of ML in Clinical Workflows}
\paragraph{Disease Prediction and Diagnosis}
The early prediction and diagnosis of diseases from medical data are one of the exciting applications of ML. Various studies have highlighted the potential of using predictive healthcare for the timely treatment of diseases. For instance, the case of cardiovascular risk prediction using different ML algorithms with clinical data is studied in \cite{weng2017can} and the study concluded that ML techniques improved the prediction efficacy. A survey of various ML techniques for the detection and diagnosis of different diseases (such as diabetes, dengue, hepatitis, heart, and liver) is presented in \cite{fatima2017survey}. The potential of using ML-based methods for prediction and prognosis of cancer is highlighted in \cite{cruz2006applications}. 

\paragraph{ML in Computer-Aided Detection or Diagnosis}
The computer-aided detection (CADe) or computer-aided diagnosis (CADx) systems are being developed mainly for the automatic interpretation of medical images that would assist the radiologist in their clinical practice. The system works by utilizing different functionalities including ML/DL, traditional computer vision and image processing techniques and relies heavily on the performance of these techniques. IBM's Watson is a classical example of CADx system developed by integrating various techniques including ML. However, any task in medical image and signal analysis automated by the application of ML/DL models can be deemed as a CADe or CADx systems, e.g., automation detection of fatty liver in ultrasound kurtosis imaging \cite{ma2016computer}. 

\paragraph{Clinical Reinforcement Learning} In reinforcement learning, the key objective is to learn a policy function for making precise decisions in an uncertain environment to maximise accumulated reward. In clinical medicine, RL can be used for providing optimal diagnosis and treatment for patients with distinct characteristics \cite{zhang2019reinforcement}. The performance evaluation of different RL techniques (i.e., Q-value iteration, tabular Q-learning, fitted Q-iteration (FQI), and deep Q-learning) for the treatment of sepsis in ICU using real-world medical dataset is presented in \cite{raghu2019reinforcement}. Sepsis is a severe infection involving organ dysfunction and is a leading cause of mortality due to expensive and suboptimal treatment. The dataset contains trajectories of a patient's physiological state and the provided treatments by clinicians at each time, along with the outcome (i.e., survival or mortality). The study concluded that simple and tabular Q-learning can learn effective policies for sepsis treatment and their performance is comparable with a complex continuous state-space method, i.e., deep Q-learning.    

\paragraph{ML for Clinical Time-Series Data}
One of the tasks in clinical workflows is the modeling of clinical time-series data. Applications of clinical time-series modeling include prediction of clinical interventions in intensive care units (ICUs) using CNN and LSTM \cite{suresh2017clinical}, mortality prediction in patients with traumatic brain injury (TBI) \cite{rau2018mortality}, and estimation of mean arterial blood pressure (ABP) and intracranial pressure (ICP) which are important indicators cerebrovascular autoregulation (CA) in TBI patients. In a recent study, attention models are used for the management of ICUs forecasting tasks (such as diagnosis, estimation, and prediction, etc.) by integrating clinical notes with multivariate and time-series measurements data \cite{song2018attend}. In a similar study, the problem of unexpected respiratory decompensation using ML techniques is investigated in \cite{ren2018predicting}. 

\paragraph{Clinical Natural Language Processing}
Clinical notes are a widely used tool by the clinicians to communicate patient state. The use of clinical text is crucial as it often contains the most important information. The progress in clinical NLP techniques is envisioned to be incorporated in future clinical software for extracting relevant information from unstructured clinical notes for improving clinical practice and research \cite{jha2011promise}. Clinical NLP offers unique challenges such as the use of acronyms, language disparity, partial structure, and quality variance, etc. The challenges and opportunities of clinical NLP for languages other than English along with a review of clinical NLP techniques is presented in \cite{neveol2018clinical}. In \cite{soysal2017clamp}, authors presented a toolkit named CLAMP that provides different state of the art NLP techniques for clinical text analysis.    

\paragraph{Clinical Speech and Audio Processing}
In the clinical environment, clinicians have to do a lot of documentation, i.e., preparing clinical notes, discharge summaries, and radiology reports, etc. According to Dr. Simon Wallace, clinicians spend 50\% of their time on clinical documentation and are highly demotivated due to clinical workload, administrative tasks, and lack of leisure time \cite{doc}. Typically, they spend more time in preparing clinical documentation as compared to interacting directly with patients. To overcome such challenges, clinical speech and audio processing offer new opportunities such as speech interfaces for interaction less services, automatic transcription of patient conversations, and synthesis of clinical notes, etc. There are many benefits for using speech and audio processing tools in the clinical environment for each stakeholder, i.e., patients (speech is a new modality for determining patient state), clinicians (efficiency and time-saving), and healthcare industry (enhance productivity and cost reduction). In the literature, speech processing has been used for the identification of disorders related to speech, e.g., vocal hyperfunction \cite{ghassemi2014learning} and as well as disorders that manifest through speech, e.g., dementia \cite{pou2018learning}. Alzheimer's disease identification using linguistic features is presented in \cite{fraser2016linguistic}. In clinical speech processing, disfluency and utterance segmentation are two well-known challenges of clinical speech processing.





\begin{figure*}
    \centering
    \includegraphics[width=0.9\textwidth]{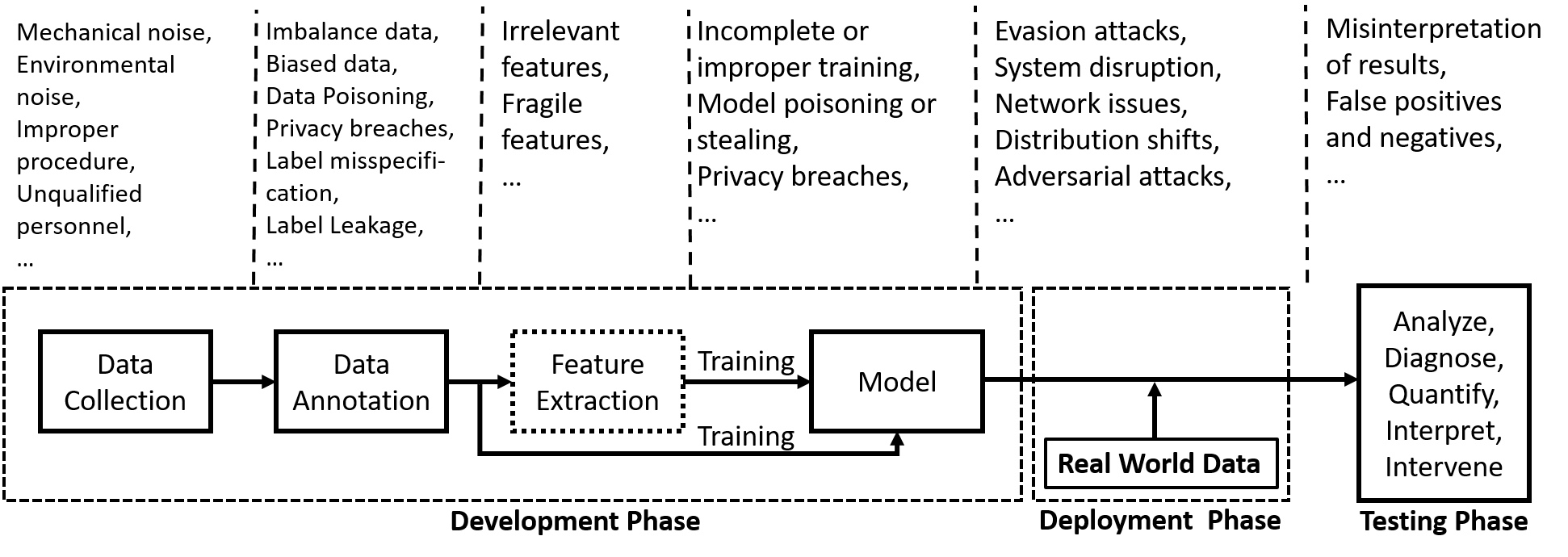}
    \caption{The pipeline for data-driven predictive clinical care and various sources of vulnerabilities at each stage.}
    \label{fig:ml_pipeline}
\end{figure*}

\section{Secure, Private, and Robust ML for Healthcare: Challenges}
\label{sec:challenges}
In this section, we analyze the security and robustness of ML/DL models in healthcare settings and present various associated challenges. 


\subsection{Sources of Vulnerabilities in ML Pipeline}
ML application in healthcare settings suffers from various privacy and security challenges that we will thoroughly discuss in this section. In addition, the three major phases of ML model development along with different potential sources of vulnerabilities causing such challenges in each step of the ML pipeline are depicted in Figure \ref{fig:ml_pipeline}.


\subsubsection{Vulnerabilities in Data Collection}
Training of ML/DL models for clinical decision support requires the collection of a large amount of data (in formats such as EHRs, medical images, radiology reports, etc.), which is in general often time-consuming and requires significant human efforts. Although in practice, medical data is carefully collected to ensure the effectiveness of the diagnosis, however, there can be many sources of vulnerabilities that can affect the proper (expected) functionality of the underlying ML/DL systems, a few of them are described next. 

\textit{Instrumental and Environmental Noise:} The collected data often contains many artifacts that arise due to instrumental and environmental disturbances. Let's consider the example of one of the widely used imagining modalities used to acquire high-resolution medical images, i.e., multishot MRI. This modality is highly sensitive to motion, and even slight movement of the subject's head or respiration can cause undesirable artifacts in the resultant image \cite{latif2018automating}, thereby increasing the risk of misdiagnosis \cite{andre2015toward}. 

\textit{Unqualified Personnel:} Healthcare ecosystems are extremely interdisciplinary and comprise of technical and non-technical personnel and often lack qualified workers that can develop and maintain ML/DL systems. As for the efficient application of data-driven healthcare, workers with strong statistical and computational backgrounds are required, e.g., engineers and data scientists. On the contrary, the clinical usability of ML/DL based systems is extremely important. Considering this aspect, hospitals tend to rely solely on physician-researchers who lack computational expertise to develop such systems \cite{manrai2014medicine}.  


\subsubsection{Vulnerabilities Due to Data Annotation}
Most applications of ML/DL in healthcare systems are supervised ML tasks which require an abundance of labelled training data. The process of assigning labels to each data sample (e.g., medical image) is known as data annotation. Ideally, this task shall mostly be performed by experienced clinicians (physicians or radiologists) to prepare domain-enriched datasets which are crucial to the development of useful ML/DL models in healthcare systems. The literature has revealed that training ML/DL models without a sound grip of the domain could be disastrous \cite{caruana2015intelligible}. However, clinicians like expert radiologists are rare professionals and hard to engage in secondary tasks like data annotation. As a result, trainee staff (with little domain expertise) or ML/DL automated algorithms are usually employed during data labelling, which often leads to many problems such as coarse-grained labels, class imbalance, label leakage, and misspecification. Some specific data annotation-based vulnerabilities are discussed as below:

\textit{Ambiguous Ground Truth:} In medical datasets, the ground truth is often ambiguous, e.g., medical image classification task \cite{finlayson2019adversarial} and even expert clinicians disagree on well-defined diagnostic tasks \cite{li2009variability}. This problem becomes more adverse with the presence of malicious users who want to perturb data, making the diagnosis difficult and causing difficulties in detecting its influence even with a human expert review.  

\textit{Improper Annotation:} The annotation of data samples process for life-critical healthcare applications should be informed by proper guidelines and various privacy and legal considerations \cite{xia2012clinical}. Most widely used healthcare datasets are annotated for coarse-grained labels whereas real-life utility of ML/DL is to highlight rare, fine-grained and hidden strata within the clinical environment. This inability to perform labelling appropriately can lead to various efficiency challenges that are discussed next.

\textit{Efficiency Challenges:}
The collections of healthcare data on which ML/DL models are built suffer from various issues that arise several efficiency challenges. A few major problems impacting the quality of data are described next. 
\begin{itemize}
    \item[(a)] \textit{Limited and Imbalanced Datasets:} The size of datasets used for training ML/DL models is not up to the required scale. In particular, one major limitation of the efficient application of DL approaches in healthcare is the unavailability of large-scale datasets, as health data is often small in size. Notably, most life-threating health conditions are naturally rare and diagnosed once in many (thousands to millions) patients. Therefore, most ML/DL algorithms can not be efficiently trained and optimized for such life-threatening healthcare task. 
    \item[(b)] \textit{Class Imbalance and Bias:} Class imbalance is yet another problem that arises in the supervised ML/DL which refers to the fact that the distribution of samples among classes is not uniform. If a class imbalanced dataset is used for training of the model then it will be reflected in the model's outcomes in terms of bias to certain categories. Biases in models' predictions in healthcare settings will have profound consequences and should, therefore, be mitigated. Various approaches have been proposed in the literature to address class imbalance problems. These approaches are discussed in the next section. 
    \item[(c)] \textit{Sparsity:} Data sparsity, i.e., missing values are common in real-world data that arise due to various reasons (e.g., unmeasured and unreported samples, etc.). Missing values and observations significantly affect the performance of ML/DL techniques. 
\end{itemize}

\subsubsection{Vulnerabilities in Model Training} 
The vulnerabilities regarding model training include improper or incomplete training, privacy breaches, model poisoning and stealing. Improper or incomplete training refers to the situations when the ML/DL model is trained with improper parameters, e.g., learning rate, epochs, batch size. 
Moreover, ML/DL models have been found strictly vulnerable to various security and privacy threats such as adversarial attacks \cite{szegedy2013intriguing}, model \cite{biggio2012poisoning} and data poisoning attacks \cite{alfeld2016data}, etc. The vulnerabilities of ML/DL systems hinder their efficient deployment for security-critical applications (such as digital forensic, bio-metrics, etc.) and as well as life-critical applications (such as self-driving cars and healthcare, etc.). Therefore, ensuring the security and integrity of the ML/DL systems is of paramount importance for such critical applications. Various security threats associated with ML/DL systems are thoroughly described in the next section.

\subsubsection{Vulnerabilities in Deployment Phase} 
The deployment of ML/DL techniques in a clinical environment essentially involves human-centric decisions. Therefore, ensuring the robustness of the system while considering fairness and accountability is necessary for the deployment phase. The following are the major vulnerabilities that can be encountered in the deployment phase of ML/DL systems. Whereas, security issues (e.g., adversarial attacks) are discussed in the next section.

\textit{Distribution Shifts:} 
Distributions shifts are very much expected in realistic healthcare settings, for example, let's consider different imaging centers and DL models trained on images of one domain (imaging center) are subsequently deployed on different domain images. In such settings, the performance of the underlying DL model degrades significantly. Moreover, in predictive healthcare, ML models are developed using historical patient data and are usually tested on the new patients which raise questions about the efficacy of the ML predictions. Moreover, such differences can be exploited for generating
adversarial examples \cite{papernot2016towards}. 

\textit{Incomplete Data:} In realistic settings, data collected for providing patient care may contain missing observations or variables, e.g., EHRs. The simplest way to avoid missing values is to ignore them completely while doing analysis but it cannot be done without knowing their relationships with already observed or unobserved data. Using the missing observations for training ML/DL models, on the other hand, leads to two well-known problems, i.e., false positives (a healthy person is diagnosed with a disease) and false negatives (a patient is identified as healthy). Both problems can have severe outcomes in actual healthcare settings, therefore, the healthcare data should be complete and compact in all aspects to ensure accurate predictions of outcomes.

\subsubsection{Vulnerabilities in Testing Phase}
Vulnerabilities in the testing phase are concerned with the interpretation of the results from the underlying ML/DL systems that include misinterpretation, false positive, and false-negative outcomes. False-positive and false-negative outcomes are due to incomplete/inefficient training of the model or due to incomplete data fed for the inference that we have discussed in the earlier section. Finally, the true essence of ML empowered healthcare is not just about turning a crank but it demands the cautious application of analytical methods \cite{pollard2019turning}.

\subsection{The Security of ML: An Overview}
In this section, we provide an overview of ML security particularly from the perspective of healthcare and highlight various associated security challenges with the use of ML.

\subsubsection{Security Threats} 
\begin{figure}[!ht]
    \centering
    \includegraphics[width=0.4\textwidth]{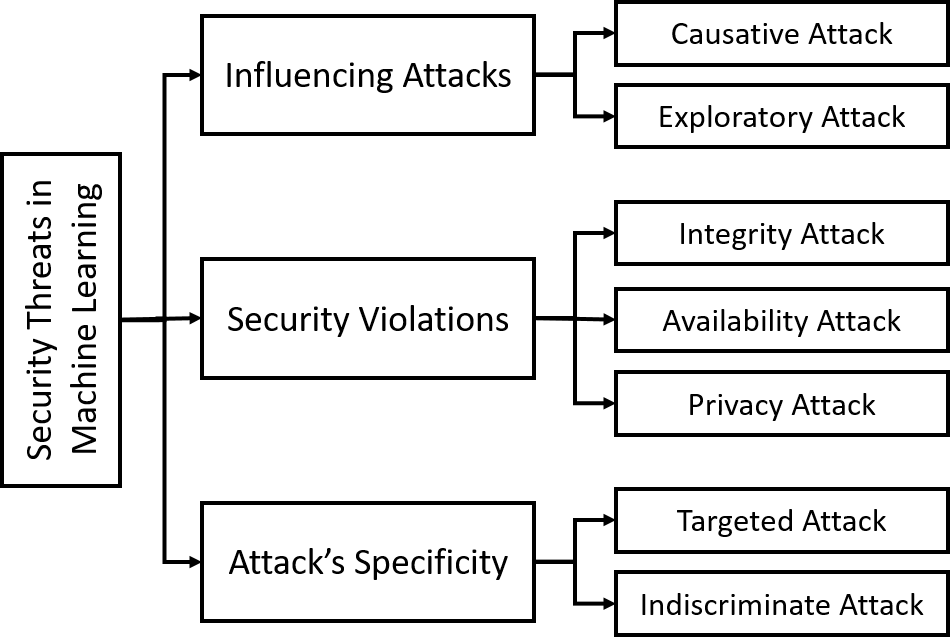}
    \caption{A taxonomy of different security threats on ML/DL models.}
    \label{fig:ml_sec_th}
\end{figure}

The security threats on ML systems can be broadly categorized into three dimensions, i.e., influence attacks, security violations, and attack specificity \cite{fredrikson2015model}. A taxonomy of these security threats on ML systems is depicted in Figure \ref{fig:ml_sec_th}.

\begin{itemize}
    \item[(a)] \textit{Influence}: Influence attacks can be of two types: (1) causative: the one that attempts to get control over training data; (2) exploratory: the one that exploits the miss-classification of the ML model without intervening the model training.
    \item[(b)] \textit{Security Violation}: It is concerned with the availability and integrity of the services and can be categorized into three types: (1) integrity attack: It attempts to increase the false-negative rate of the deployed model (classifier) when the model is given harmful inputs; (2) availability attack: Unlike integrity attack, it tries to achieve an increase in the false-positive rate of the classifier in response to benign inputs; (3) privacy violation attack: It is concerned with the unveiling of sensitive and confidential information of the training data, trained model or both. 
    \item[(c)] \textit{Attack Specificity}: The specificity of an attack can be defined in two ways: (1) targeted attack: whether the attack is intended for a specific input sample or a group of samples; (2) indiscriminate attack: it causes the ML model to fail indiscriminately. 
\end{itemize}

\subsubsection{Adversarial Machine Learning (ML)}
Adversarial attacks are the result of recent efforts for identifying vulnerabilities in ML/DL models training and inference. Adversarial attacks have appeared as one of the biggest security threats to ML/DL systems \cite{szegedy2013intriguing,goodfellow2014explaining,papernot2016limitations,papernot2017practical,usama2019adversarial}. In adversarial attacks, the key goal of an adversary is to generate adversarial examples by adding small carefully crafted (unnoticeable) perturbation into the actual (non-modified) input samples to evade the integrity of the ML/DL system. In general, there are two types of adversarial attacks that are described next. 

\begin{enumerate}
    \item[(a)] \textit{Poisoning Attacks: }Adversarial attacks affecting the model training, i.e., manipulating the training data to mislead the learning of ML/DL model are known as poisoning attacks \cite{biggio2012poisoning}. 
    \item[(b)] \textit{Evasion Attacks:} Adversarial attacks on the inference phase of the training process are known as evasion attacks \cite{biggio2013evasion}. In such attacks, an attacker manipulates the test data to compromise the integrity of the ML/DL model to harmful inputs. 
\end{enumerate}


In healthcare applications, poisoning attacks are highly relevant because direct manipulation of the training data may be difficult or even impossible in some cases. Alternatively, the addition of new samples might be relatively easy, however, any such consequences hinder the applicability of the ML/DL systems. Therefore, the detection of poisoning attacks is critical for the robust application of ML/DL in healthcare applications. For instance, systematic poisoning attacks against six conventional ML models that were developed for hypothyroid diagnosis are presented in \cite{mozaffari2014systematic}, where the objective of the attacker was to prevent hypothyroid diagnosis. 

Similarly, a few researchers have highlighted the threat of these attacks to ML/DL models in healthcare settings and we provide insights from such articles in this section. Unlike adversarial examples created for evading ML/DL models in other settings, the concept of \textit{adversarial patients} for healthcare applications is introduced in \cite{papangelou2018toward}. The authors argue that rather than intentional adversarial examples, the caution should be for unintentional adversarial patients that can lead to severe ethical issues. They identified a subgroup of adversarial patients and empirically validated that patients with identical predictive features can have significantly
different individual treatment effects. In recent studies, white box and black box adversarial attacks have been demonstrated against three clinical applications; namely, fundoscopy, dermoscopy, and chest X-ray analysis \cite{finlayson2019adversarial,finlayson2018adversarial}. Furthermore, in \cite{finlayson2018adversarial}, authors highlighted various potential incentives for adversaries via adversarial attacks in clinical trials that will rise with the increasing use of ML in the future, particularly, with the emergence of computer-aided diagnosis and decision support systems.     


\subsection{ML for Healthcare: Challenges}
In this section, we discuss various challenges which hinders the applicability of ML/DL systems in practical healthcare applications. 

\subsubsection{Safety Challenges}: 
Excellent performance in a controlled lab environment (which is a common ML community practice) is not evidence of safety. Safety of ML/DL is the determination of how safe the ML/DL system is for patients. There should be a constant thought of safety throughout the ML/DL lifecycle. Majority of routine clinicians tasks are mundane, and patients they encounter have common health conditions. It is their role of diagnosing rare, subtle, and hidden health conditions which occur once in millions. Enabling ML/DL to performing well on hidden strata, outliers, edge, and subtle cases is key to ensure the safety of current AI systems. 


\subsubsection{Privacy Challenges}
Privacy is one of the major challenges in data-driven healthcare which is concerned with the use of users' data by the ML/DL systems for making predictions. The users (i.e., patients) expect that their healthcare service providers are following necessary safety measures to safeguard their inherent right to the privacy of their confidential information, e.g., age, sex, date of birth, and health data. Potential privacy threats can be of two types, i.e., unveiling confidential information and malicious use of data (potentially by unauthorized agents). 

Privacy depends upon the characteristics and nature of the data being collected, the environment it has been created in, and patients' demographics. Therefore, mitigation of privacy breaches using the appropriate technique(s) is critical as such breaches can directly harm the patients. The confidential data should be anonymized to prevent privacy breaches such as (re-)identification of the individuals \cite{al2019privacy}. Moreover, necessary attention should be paid to understand privacy concerns at each stage of data processing and the transfer of data among different departments within a hospital should be communicated in a secure environment. 

\subsubsection{Ethical Challenges}
In user-centric applications of ML such as healthcare, it is important to ensure the ethical use of data. Explicit measures should be taken to understand the targeted user population and their sociological aspects before collecting data for building ML models. Moreover, understanding how data collection can harm a patient's well-being and dignity is an important consideration in this regard. If ethical concerns are not taken into account then the application of ML in realistic settings will have adverse results. Furthermore, to ensure fair and ethical operation of automated systems, it is imperative to have a clear understanding of the AI system in uncertain and complex scenarios \cite{zhang2018fairness}. 



\subsubsection{Causality is Challenging}
Understanding causality is important in healthcare because most of the crucial healthcare problems require causal reasoning, i.e., \textit{``what if?''} \cite{schulam2017reliable}.  For example, asking a question about what will happen if a doctor prescribed treatment $A$ instead of treatment $B$. Such questions cannot be exploited through classical learning algorithms and to answer them we need to analyze the data from the lens of causality \cite{ghassemi2018opportunities}. In healthcare, learning is often solely based on observational data and asking causal questions by learning from observational data is quite challenging which requires building causal models.

DL models are black-box which lacks fundamental underlying theory and these models essentially work by exploiting patterns and correlations without considering any causal link \cite{begoli2019need}. In general, this cannot be deemed as a limitation since prediction does not require any causal relation. In predictive healthcare, the absence of causal relation can raise questions about the conclusions that can be drawn from outcomes of DL models. Furthermore, fairness in decision making can better be enforced through the lens of causal reasoning \cite{khademi2019fairness,kilbertus2017avoiding}. The estimation of the causal effect of some variable(s) on a target output (e.g., target class in multi-class classification problem) is important to ensure fair predictions.

\subsubsection{Regulatory and Policy Challenges}
The full potential of ML/DL systems (which essentially constitutes software as a medical device) in actual healthcare settings can only be realized by addressing regulatory and policy challenges. The literature suggests that the regulatory guidelines are needed for both medical ML/DL systems and their integration in actual clinical settings \cite{faes2019automated}. Therefore, the integration of AI-empowered ML/DL systems in the actual clinical environment should be in compliance with the policies and regulations defined by the government and regulatory agencies. However, existing regulations are not suitable for certifying systems which are ever-evolving such as ML/DL empowered systems because yet another key challenge with the use of ML/DL algorithms in clinical practice is to determine how these models should be implemented and regulated since these models will incorporate learning from the new patient data \cite{ochallenges}. In addition, the objective clinical evaluation of ML/DL systems for particular clinical settings is crucial to ensure safe, effective, and robust operation that does not harm the patients in either way. Data scientist and AI engineers should be employed in hospitals for assessing AI systems regularly to ensure it is still safe, relevant, and working fine.

\begin{table*}[]
\centering
\caption{Summary of the state of the art data security and privacy preserving methods in healthcare settings.}
\scalebox{0.9}{
\begin{tabular}{|l|l|p{40mm}|l|p{30mm}|}
\hline
\textbf{Authors} & \textbf{Goal} & \textbf{Method} & \textbf{ML Model(s)} & \textbf{Medical Dataset(s)} \\ \hline
David et al. \cite{david2015efficient} & \multirow{8}{*}{Privacy} & Commodity based cryptography. & \begin{tabular}[c]{@{}l@{}}Hyperplane decision\\ and Naive Bayes classifiers.\end{tabular} & N/A \\ \cline{1-1} \cline{3-5} 
Zhu et al. \cite{zech2018natural} &  & Polynomial aggregation and multiparty random masking. & SVM with nonlinear kernel. & N/A \\ \cline{1-1} \cline{3-5} 
Jagielski et al. \cite{jagielski2018manipulating} &  & Proposed an algorithm names as TRIM to defend poisoning attacks. &  Linear Regression  &  Anticoagulant drug
Warfarin  \\ \cline{1-1} \cline{3-5} 
Liu et al. \cite{liu2016collaborative} &  & XMPP server and several mobile  devices.  & Proposed a DL framework, & Human Activity Recognition \\ \cline{1-1} \cline{3-5} 
Malathi et al. \cite{malathi2019hybrid} &  & Paillier homomorphic encryption.  & NaïveBayesia, SVM, NeuralNetwork, and FKNN--CBR & Indian Liver Patient \\ \cline{1-1} \cline{3-5} 
Takabi et al. \cite{takabi2016privacy} &  & Homomorphic encryption.  & DNN  & 15 datasets from UCI repository. \\ \hline
Kim et al. \cite{kim2018secure} & \multirow{1}{*}{Security} & Homomorphic encryption based secure logistic regression. & Logistic Regression & Five medical datasets having binary classes. \\ \hline
\end{tabular}}
\label{tab:summary}
\end{table*}

\subsubsection{Availability of Good Quality Data}

The availability of representative, diverse and high-quality data is one of the major challenges in healthcare. For instance, the amount of data available to the research community is very small in size and limited in scope as compared to the heterogeneous collections of large-scale multi-modal patient data being generated on daily basis by different small and large size healthcare institutions. However, the development of good quality data that resembles real clinical settings is on the other very challenging and requires resources for management and maintenance. The availability of high-quality data can effectively serve the intended purpose of disease prediction and decision making for planning treatment. 


The data collected in practice suffer from different issues such as subjectivity, redundancy, and bias. As the ML/DL models perform inferences by solely learning the latent factors of the data on which they are trained, therefore, the effect of data generated by the undesirable past practices of hospitals will be reflected in the outcomes of the algorithm. For example, most people with no health insurance are denied healthcare services and if AI learns from that data, it will do the same. It has been shown that a model could depict racial bias by producing varying outcomes for different subpopulations \cite{chen2018my} and the training data can also introduce its own modeling challenges \cite{ghassemi2019practical,panch2019inconvenient}.


\subsubsection{Lack of Data Standardization and Exchange}
Medical ML/DL system shall facilitate a deep understanding of the underlying healthcare task, which  (in most cases) can only be achieved by utilising other forms of patients data. For example, radiology is not all about clinical imaging. Other patient EMR data is crucial for radiologists to derive the precise conclusion for an imaging study. This calls for the integration and data exchange between all healthcare systems. Despite extensive research on data exchange standards for healthcare, there is a huge ignorance in following those standards in healthcare IT systems which broadly affects the quality and efficacy of healthcare data, accumulated through these systems. There are numerous guidelines to perform specific medical interventions like imaging studies (i.e., with define exposure and positioning) to ensure the significance of the data clinically. However, current healthcare IT systems largely ignore standards and clinicians barely follow well-established guidelines. As a result, data integration and exchange efforts across different specialities and organisations fail. Data integration to match diverse patients' medical records is crucial to deliver high-value patient care. The lack of appetite to implement data exchange standards in wider healthcare industry hinders the efficacy of ML/DL systems as multi-modal data is vital to ensure the deep understanding of algorithms, and will undoubtedly enhance the performance of physicians towards clinical decisions using data driven insights.


\subsubsection{Distribution Shifts}
The problem of data distribution shifts is yet another major challenge and perhaps one of the most challenging problems to solve \cite{perone2019unsupervised}. In clinical practice, training and testing data distributions can diverge due to many reasons, e.g., medical data is generated by different institutions using different devices for patients having complicated cases. Due to this issue, ML/DL models developed using available public databases (by the scientific community and academicians) do not give expected performance when deployed in an actual clinical environment. Distribution shifts are frequent in the medical domain, in particular, medical imagining where different protocols and parameter choices can result in images of significantly different distributions. ML models are typically trained under the principle of empirical risk minimization (ERM) which provides good learning bounds and guarantees if its assumptions are satisfied. For instance, one of the foremost and strong assumptions is that both the training and test datasets are derived from a similar domain (i.e., data distributions). However, this assumption is not valid in practice, and models trained under such an assumption fail to generalize to other domains In contrast, the life-critical nature of clinical applications demands a smooth and safe operation of ML/DL techniques.

\subsubsection{Updating Hospital Infrastructure is Hard}
Healthcare IT systems are mostly proprietary and operate in silos, which results in the revision, fixing, and update of software being costly and time-consuming. It has been reported in the literature that in 2013, the majority of hospitals were using the ninth version of the international classification of disease (ICD) system---even though a revised version (i.e., ICD-10) was released as early as 1990 \cite{finlayson2019adversarial}. The difficulties in updating hospital software infrastructure can raise many vulnerabilities with the use of modern tools like ML/DL systems.

\begin{figure*}[!ht]
    \centering
    \includegraphics[width=0.7\textwidth]{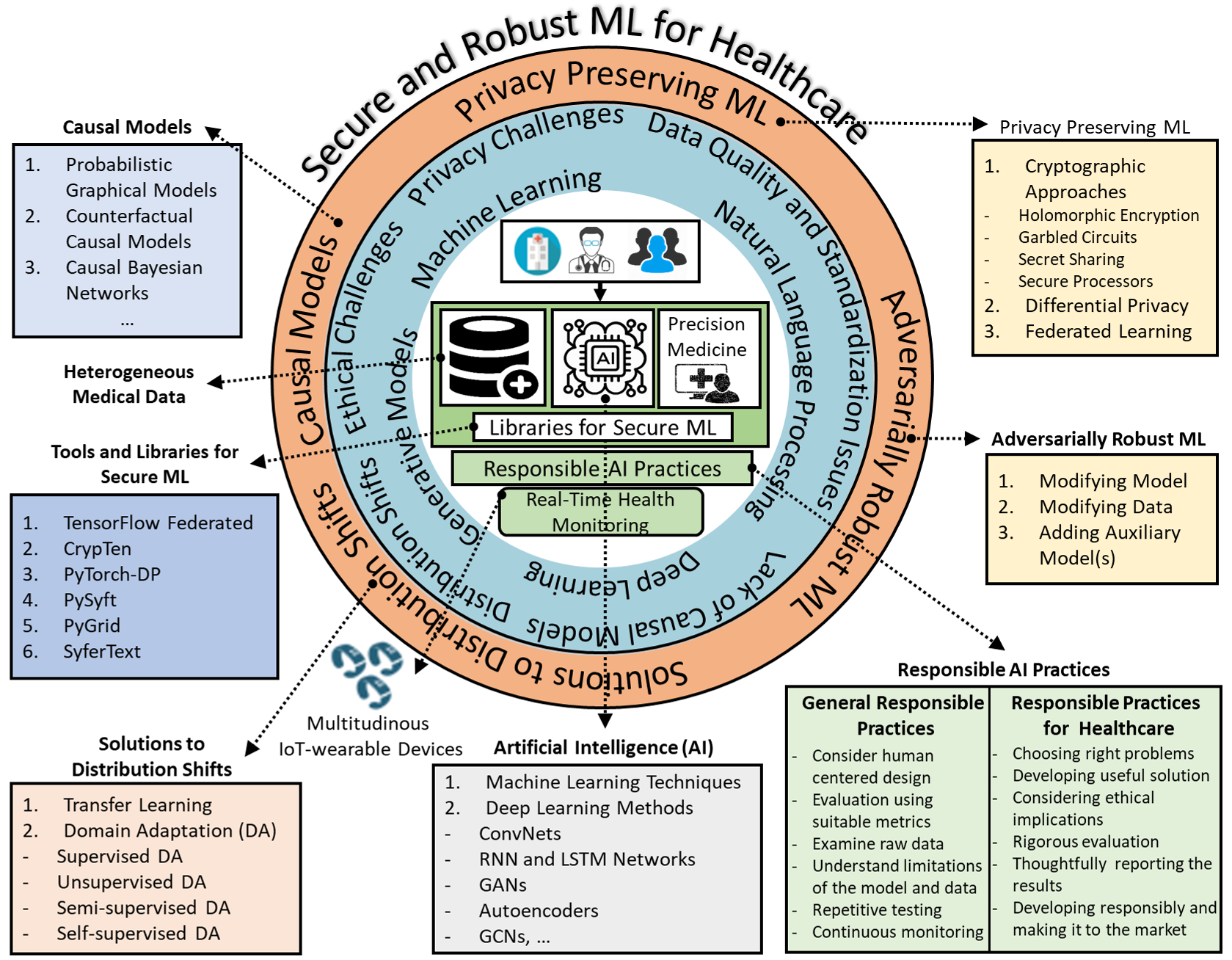}
    \caption{A taxonomy of commonly used approaches for secure, private, and robust ML.}
    \label{fig:secure_robust_ml}
\end{figure*}

\section{Secure, Private, and Robust ML for Healthcare: Solutions}
\label{sec:solutions}
  
In this section, we present an overview of various proposed methods to ensure secure, private, and robust ML for healthcare applications. A summary of articles focused on the topic of ``secure and privacy-preserving ML for healthcare'' is presented in Table \ref{tab:summary} and various approaches for secure, private, and robust ML are described next. In addition, a taxonomy of commonly used approaches for secure, private, and robust ML is presented in Figure \ref{fig:secure_robust_ml} and described individually next. 

\subsection{Privacy-Preserving ML}
Preserving the privacy of the user in healthcare is paramount, as it is a user-centric application and involves the collection of personal data and any breach of privacy can lead to unavoidable consequences. Preserving privacy means that ML model training and inference should not reveal any additional information about the subjects from whom data was collected. In general, ML/DL requires training data stored on a central repository (e.g., cloud) that may include the users' private data which raises various threats and to address such concerns data anonymization techniques are used. However, it has been reported in the literature that meaningful information can be inferred about individuals' private data even when the data is anonymized \cite{narayanan2008robust}.

Various efforts in the literature have addressed the privacy issues with the use of ML. Three different protocols for the two-server model are presented in \cite{mohassel2017secureml}, where the private data is distributed among two non-colluding servers by the data owners and then those servers train the ML models on the joint data by following secure two-party computation (2PC). Furthermore, different techniques have been proposed to perform secure arithmetic operations in the secure multi-party computational environment and alternatives to nonlinear activation functions used in ML models such as softmax and sigmoid are also proposed. Similarly, various techniques for privacy-preserving ML such as cryptographic and differential privacy approaches are discussed in \cite{al2019privacy}. Here we briefly discuss the widely used methods for preserving privacy.    

\subsubsection{Cryptographic Approaches}
Cryptographic approaches are used in the scenarios where the ML model requires encrypted data (for training and testing purposes) from multiple parties. The widely used methods include homomorphic encryption, secret sharing, garbled circuits, and secure processors which are briefly described next.  

\begin{itemize}
    \item[(a)] \textit{Homomorphic Encryption:} It enables computations on encrypted data with operations such as addition and multiplication which can be used as a basis for computing complex functions. Typically, the data is encrypted using ciphertext and public keys of the original data owners.  
    
    \item[(b)] \textit{Garbled Circuits:} Garbled circuits are used in cases where two parties (let's assume Alice and Bob) want to get results computed using their private data. Alice will send the function in the form of the garbled circuit along with her input. After obtaining the garbled version of his input from Alice in oblivious fashion, Bob will use his garbled input with the garbled circuit to get the result of the required function and can share it with Alice, if required. The use of homomorphic encryption and garbled circuits to build cryptographic blocks for developing three classification techniques; namely, Naïve Bayes, decision trees, and hyperplane decision is presented in \cite{bost2015machine}, where the goal is to protect ML models and new samples submitted for inference. 
    
    \item[(c)] \textit{Secret Sharing:} The strategy of distributing secrete among multiple parties while holding a ``share'' of the secret is known as secret sharing. The secret can only be reconstructed when all individual shares are combined; otherwise, they are unuseful. In some settings, the secret is reconstructed using $t$ shares (where $t$ is a threshold value) that will not require all shares to be combined. A secret sharing paradigm for computing privacy-preserving parallelized principal component analysis (PCA) is presented in \cite{bogdanov2018implementation}. In a similar study \cite{bonawitz2017practical}, a protocol is developed using the ``secret sharing'' strategy for aggregating model updates received from multiple input parties, the updates are used for training of the ML model. A privacy-preserving emotion recognition framework is presented in \cite{hossain2019emotion}. Authors used a multi-secret sharing scheme for transmitting audio-visual data collected from users using edge devices to the cloud where a CNN and sparse autoencoder were applied for feature extraction and support vector machine (SVM) was used for emotion recognition. 
    
    \item[(d)] \textit{Secure Processors:} Secure processors were originally developed by rogue software to ensure the confidentiality and integrity of sensitive code from unauthorized access at higher privilege levels. However, these processors are being utilized in privacy-preserving computation, e.g., Intel SGXprocessor. For instance, Ohrimenko et al. developed an SGX-processor-based data oblivious system for k-mean clustering, decision trees, SVM, and matrix factorization \cite{ohrimenko2016oblivious}. The key idea was to enable collaboration between multiple data owners running the ML task on an SGX-enabled data center. All types of communications between the data owners and the enclave were performed by establishing independently a secure channel (i.e., an individual channel for each data owner). 
\end{itemize}

\subsubsection{Differential Privacy}

Differential privacy refers to the mechanism of adding perturbation into the datasets to protect private data. The idea of adding adequate noise in the database for preserving privacy was first introduced by  C. Dwork in 2006 \cite{dwork2011differential}. Differential privacy constitutes a strong standard for guaranteeing privacy for algorithms performing analysis on aggregate databases and it is defined in terms of the application-specific concept of neighbor datasets \cite{abadi2016deep}. Differential privacy is particularly useful for applications like healthcare due to its several properties such as group privacy, composability, and robustness to auxiliary information. Group privacy implies elegant degradation of privacy guarantees when datasets contain correlated samples. Whereas, composability enables modularity of the algorithmic design, i.e., when individual components are differentially private. Robustness to auxiliary information means that the privacy of the system will not be affected by the use of any side's information that is known to the adversary. To avoid privacy breaches, the researchers can also explore encrypted and noisy datasets for building ML empowered healthcare applications \cite{mcdermott2019reproducibility}.

Various approaches for differential privacy have been proposed in the literature, e.g., private aggregation of teacher ensembles (PATE) for private ML \cite{papernot2018scalable}, differentially private stochastic gradient descent (DP-SGD) algorithm \cite{abadi2016deep}, moments accountant \cite{wang2018subsampled}, hyperparameter selection \cite{mcmahan2018general}, Laplace \cite{phan2017adaptive} and exponential noise differential privacy mechanisms \cite{mcsherry2007mechanism,dwork2009exponential}. For instance, privacy-preserving distributed DL for clinical data using differential privacy that incorporates the idea of cyclical weight transfer is presented in \cite{beaulieu2018privacy}.

\subsubsection{Federated Learning}


The idea of federated learning (FL) has been recently proposed by Google Inc. \cite{mcmahan2016communication}. In FL, a shared ML model is built using distributed data from multiple devices where each device trains the model using its local data and then shares the model parameters with the central model without sharing its actual data. An FL-based decentralized scheme using iterative cluster primal-dual splitting (cPDS) algorithm to predict hospitalization requiring patients using large-scale EHR of heart-related diseases is presented in \cite{brisimi2018federated}. In \cite{vepakomma2018split}, simple vanilla, U-shaped, and vertically partitioned data-based configurations for split learning DL models are presented. The proposed framework is named SplitNN that does not require sharing of patients' critical data with the server. A framework of federated autonomous deep learning (FADL) using distributed EHR is presented in \cite{liu2018fadl}.

\begin{figure*}
    \centering
    \includegraphics[width=0.9\textwidth]{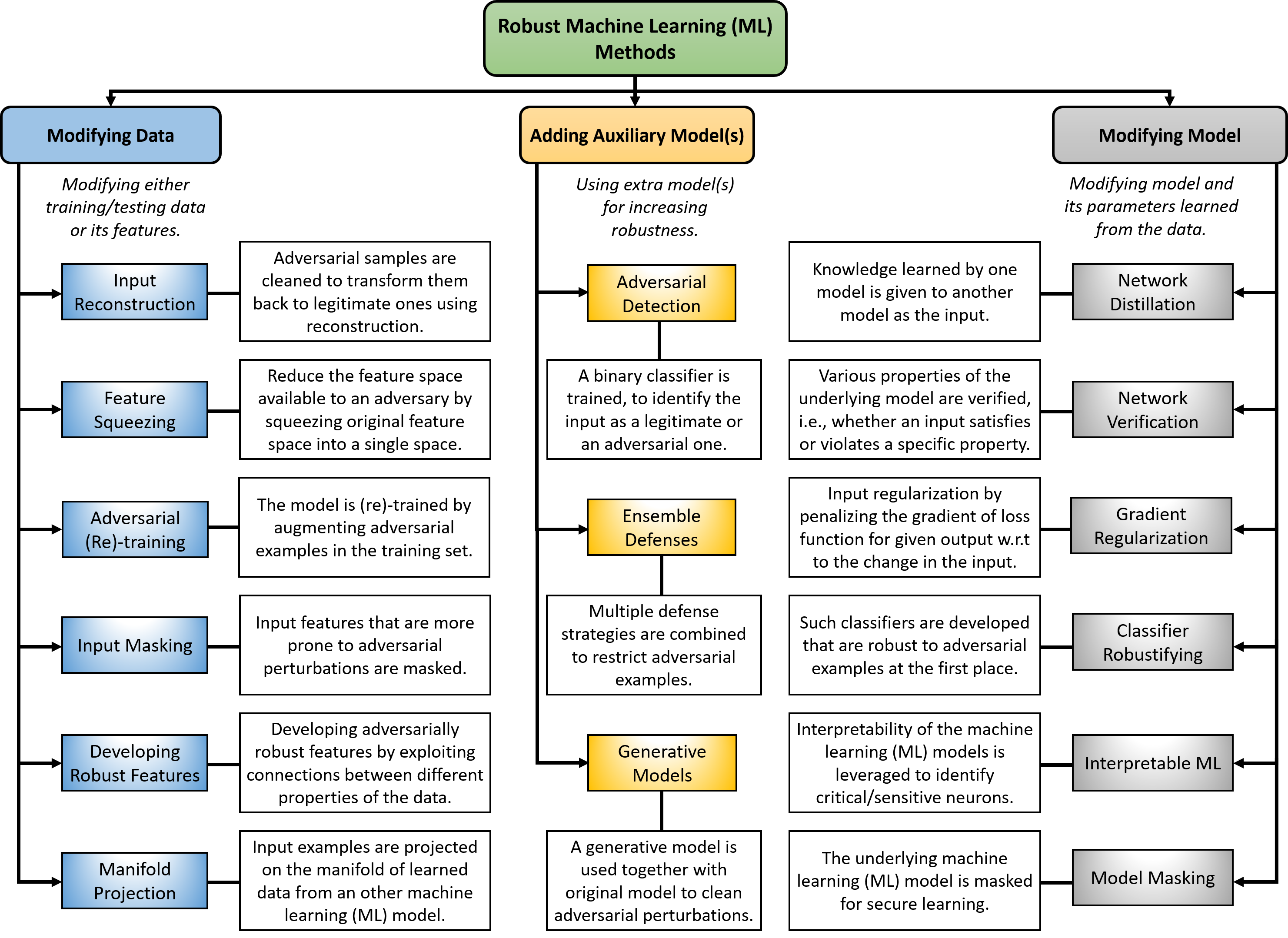}
    \caption{Taxonomy of Adversarial Defenses (Source: \cite{qayyum2019securing}). Defenses are categorized into three categories: (1) Modifying Data; (2) Modifying Model; and (3) Adding Auxiliary Model(s).}
    \label{fig:taxonomy_defenses}
\end{figure*}

\subsection{Countermeasures Against Adversarial Attacks} 
In the recent literature, countermeasures against adversarial attacks are categorized into three classes: (1) modifying model; (2) modifying data; and (3) adding an auxiliary model(s) \cite{qayyum2019securing}. A taxonomy of such methods is presented in Figure \ref{fig:taxonomy_defenses} and are discussed next.  

\subsubsection{Modifying Model}
The modifying model includes methods that modify the parameters or features of the trained ML model, widely used methods include the following: 
\begin{itemize}
    \item \textit{Defensive Distillation:} The distillation of neural networks was first introduced by Hinton et al. as a method for transferring the knowledge from a larger model to a smaller one \cite{hinton2015distilling}. The notion of network distillation was then adopted by Papernot et al. to defend against adversarial attacks, also known as defensive distillation \cite{papernot2016distillation}. The authors used the predicted labels of the first model as the labels of the input sample to the original DL model. This strategy increases the robustness of the DL model to considerably small perturbations. However, in a later study, Carlini and Wagner demonstrated that their proposed adversarial attack (named as C\&W attack) evaded the defensive distillation method \cite{carlini2017adversarial}. 
    
    \item \textit{Network Verification:} The techniques verifying certain properties of DL models in response to input samples are known as network verification methods. The key goal is to restrain adversarial examples while checking whether the input satisfied or violated certain properties. In \cite{katz2017reluplex}, such a method is proposed that uses ReLU activation and satisfiability modulo theory (SMT) to make deep models resilient against adversarial attacks.  
    
    \item \textit{Gradient Regularization:} The idea of using input gradient regularization for defending adversarial examples was proposed by Ross et al.\cite{ross2018improving}. They trained the differentiable models by regularizing the variation in the results with respect to the change in the input due to which small adversarial perturbations were not able to affect the output of DL models. However, this method increases the complexity of the training process by a factor of two. 
    
    \item \textit{Classifier Robustifying:} In this method, classification models are developed that are robust to adversarial attacks rather than building a detection strategy for such attacks. In \cite{bradshaw2017adversarial}, authors exploited the uncertainty around the adversarial examples and proposed a hybrid model by utilizing Gaussian processes (GPs) with RBF kernels on top of DNNs to make them robust against adversarial attacks. In a similar study, a robust model is proposed for MNIST classification that uses analysis by synthesis through learned class-conditional data distribution. 
    
    \item \textit{Interpretable ML:} It includes those methods that aim at explaining and interpreting the outcomes of ML/DL models for robustifying them against adversarial attacks. An approach utilizing the interpretability of deep models for the detection of adversarial examples for face recognition task is presented in a recent study \cite{tao2018attacks}. The key aspect of this method is that it identifies critical neurons for the individual task by initiating a bi-directional correspondence reasoning between the model's parameters and its attributes. The activation values of the identified neurons are then increased to augment the reasoning part and activation values of other neurons are decreased to mask the uninterpretable part. However, Nicholas Carlini demonstrated that the aforementioned method utilizing the interpretability of deep models is not resilient to untargeted adversarial examples generated using $L_\infty$ norm \cite{carlini2019ami}.
    
    \item \textit{Masking ML Model:} In a recent study \cite{nguyen2018learning}, a method for secure learning is presented in which the problem of adversarial ML is formulated as learning and masking problem. The masking of the deep model was performed by introducing noise in the \textit{logit} output which successfully deafened attacks with low distortions. 
\end{itemize}

\subsubsection{Modifying Data}
It includes those methods that aim at either modifying the data or its features, commonly used methods are described next: 

\begin{itemize}
    \item \textit{Adversarial (Re-)training:} This is a very basic method that was originally proposed by Goodfellow et al. for making deep models robust to adversarial examples \cite{goodfellow2014explaining}. In this method, the ML/DL models are trained (or retrained) using an augmented training set that includes adversarial examples. Various studies have used this method for evaluating the robustness of DL classifiers using different datasets, e.g., MNIST \cite{huang2015learning} and ImageNet \cite{katz2017reluplex}. However, it has been reported in the literature that this method fails to defend against iterative adversarial perturbation generation methods like basic iterative method (BIM) \cite{kurakin2016adversarial}. 
    
    \item \textit{Input Reconstruction:} The method of transforming adversarial examples into legitimate ones by cleaning the adversarial noise is known as input reconstruction. The transformed samples have no harmful effect on the inference of deep models. In  \cite{gu2014towards}, denoising autoencoder is used for the cleaning of adversarial examples.   
    
    \item \textit{Feature Squeezing:} Xu et al. \cite{xu2017feature} proposed feature squeezing as a defense method against adversarial examples by squeezing the input feature space that an adversary can exploit to construct adversarial examples. To reduce the available feature space to an adversary, authors combined heterogeneous feature vectors in the original feature space into a single space. The feature squeezing was performed at two levels: (1) smoothing the spatial domain using local and non-local operations and (2) minimizing color bit depth. Moreover, the performance evaluation of the proposed defense was performed using eleven state of the art adversarial perturbation generation methods using three benchmark datasets (i.e., CIFAR10, MNIST, and ImageNet). However, in a later study, the aforementioned defense method was found to be less effective \cite{he2017adversarial}. 
    \item \textit{Features Masking:} The method of feature masking was proposed by Gao et al. \cite{gao2017deepcloak} that aims at masking the most sensitive features of the input that are susceptible to adversarial perturbations. The authors added a masking layer right before the classification layer (i.e., softmax) that sets the corresponding weights of the sensitive neurons to zero. 
    
    \item \textit{Developing Adversarially Robust Features:} To develop adversarially robust features, the connections between the metric of interest and natural spectral geometrical property of the dataset has been leveraged in \cite{garg2018spectral}. Furthermore, the authors provided empirical evidence about the effectiveness of using a spectral approach for developing adversarially robust features.  
    
    \item \textit{Manifold Projection:} The method of projecting input samples on the manifold learned by the generative models is known as manifold projection. Song et al. \cite{song2017pixeldefend} used generative models to clean adversarial noise (perturbations) from the adversarial images then the cleaned images are used as the input to the non-modified model.  In a similar study \cite{shen2017ape}, generative adversarial networks (GANs) are used for cleaning of adversarial noise. 
    
\end{itemize}

\subsubsection{Adding Auxiliary Model(s)}
In these methods, additional auxiliary ML/DL models are integrated to robustify the mainstream model, commonly used methods that fall into this class are described in the following paragraphs:  

\begin{itemize}
    \item \textit{Adversarial Detection:} In this method, an additional binary classifier is trained to distinguish between the adversarial and original samples that can be regarded as the detector model \cite{lu2017safetynet,gopinath2017deepsafe}. In \cite{metzen2017detecting}, a simple DNN based detector model is used for the detection of adversarial examples. Similarly, an outlier class has been introduced during the training of a deep model that helps the model to detect the adversarial examples belonging to the outlier class. 
    
    \item \textit{Ensembling Defenses:} The literature suggests that adversarial examples can be constructed in multi-faceted fashion. Therefore, to develop an efficient defense method against such adversarial examples, multiple defense strategies can be integrated sequentially or in parallel \cite{kurakin2018ensemble}. The PixelDefend method is an excellent example of an ensemble defense method in which authors used an ensemble of two methods, i.e., adversarial detection and input reconstruction \cite{song2017pixeldefend}. However, it has been shown that the ensemble of weak defenses does not necessarily increase the robustness of DL models to adversarial attacks \cite{he2017adversarial}.  
    
    \item \textit{Using Generative ML Models:} The idea of defending against adversarial attacks by utilizing generative models was firstly presented by Goodfellow et al. \cite{goodfellow2014explaining}, however, in the same study the authors presented an alternative hypothesis of ensemble training and articulated that generative training is not sufficient. In \cite{santhanam2018defending}, adversarial examples are cleaned using GAN that was trained on the same dataset. In a similar study \cite{samangouei2018defense}, a framework named Defense-GAN is presented that is trained on the distribution of legitimate samples. Defense-GAN finds similar output during the testing phase without adversarial perturbations that are given as input to the original DL model. 
\end{itemize}

\subsection{Causal Models for Healthcare}
Asking causal questions in healthcare is a very challenging yet important approach and ideally, causal inferences require experiments. But it in healthcare this not always possible, e.g., if we want to figure out what will happen if a person takes drug $A$ instead of $B$, we can not experiment it directly on the patient which is unethical and can have unintended consequences. Alternatively, retrospective observational data is leveraged to train models for making counterfactual predictions of what we would have observed if we had run an experiment \cite{schulam2017if}. Causality can be deemed in two foundational ways, i.e., potential outcomes and causal graphical models that require manipulating reality. In predictive healthcare, potential outcomes can be treatment, action, and interventions. If the total number of possible treatments is $T$ then we can have $T$ possible outcomes and the unit of observation will be a patient who gets one of the $T$ treatments. 

In the literature, different approaches have been presented for providing causal inferences and reasoning in healthcare using classical models. For instance, the Gaussian processes based counterfactual causal model has been presented in \cite{schulam2017if} and in a similar study, authors introduced the counterfactual Gaussian process (CGP) for predicting counterfactual future progression and argued that counterfactual model can provide reliable decision support \cite{schulam2017reliable}. The use of probabilistic graphical models to analyze causality in health conditions for identification sleep apnea, Alzheimer's disease, and heart diseases is presented in \cite{sato2015probabilistic}. A comprehensive review of graphical causal models can be found in this recent study \cite{glymour2019review}.

\subsection{Solutions to Address Distribution Shifts}
To cater with data distribution shift problem various techniques have been proposed in the literature (e.g., transfer learning and domain adaptation), which are described next. 

\subsubsection{Transfer Learning}
The requirement of the availability of a large-scale dataset for training DL models capable of providing high performances can be partially mitigated using transfer learning. Transfer learning is a technique in which a model trained on a larger dataset is re-trained (fine-tuned) on the application-specific dataset (relatively smaller in size to the first one). The aim is to transfer knowledge learned by the model from one domain (data distribution) to the other domain \cite{yosinski2014transferable}. However, transfer learning can be problematic for healthcare applications due to the requirement of sufficiently large data for first training and good quality data annotated by expert clinicians such as radiologists for domain-specific training. 

\subsubsection{Domain Adaptation}
Domain adaptation is the method of learning a DL model by considering a shift between the training (often called as source domain) and test (often called as target domain) data distributions, i.e., source domain and target domain distributions are different. Domain adaptation is a special case of transfer learning that can be particularly useful for medical image analysis tasks such as MRI segmentation \cite{perone2019unsupervised,ghafoorian2017transfer}, chest X-ray classification \cite{madani2018semi}, and multi-class Alzheimer disease classification \cite{wachinger2016domain}, etc. Different facets of domain adaptation have been proposed in the literature and can be broadly categorized as supervised, unsupervised, semi-supervised, and self-supervised domain adaptation methods which are described below. Please note that the definition of domain adaptation is ambiguous since it may refer to labeled data being available in the source or target domains and the definitions provided below for each method are mostly used in the literature \cite{wilson2019survey}.   

\begin{itemize}
    \item[(a)] \textit{Supervised Domain Adaptation:} This method is similar to a supervised learning strategy with the only difference of different distributions for source domain and target domain data. Supervised domain adaptation is particularly useful when a labeled data is available for the target domain and generally, the source domain also has labeled data. 
    \item[(b)] \textit{Unsupervised Domain Adaptation:} In unsupervised domain adaptation, source domain data is labeled and target domain data is unlabeled. An unsupervised domain adaptation method using reverse flow and adversarial training for generating synthetic medical images is presented in \cite{mahmood2018unsupervised}. In addition, the authors used self-regularization for preserving clinically-relevant features. 
    \item[(c)] \textit{Semi-supervised Domain Adaptation:} In semi-supervised domain adaptation, labeled source data and partial labeled target domain. 
    \item[(d)] \textit{Self-supervised Domain Adaptation:} Self-supervised domain adaptation methods aims at learning visual models without manual labeling by training generic models using auxiliary relatively simple tasks (known as pretext tasks). The supervision is provided by modifying the original visual content (e.g., a set of images) according to known transformations (e.g., rotation) and then the model is trained to predict such transformations that serve as labels for the pretext tasks \cite{xu2019self}. 
\end{itemize}

\subsection{Towards Responsible ML}
In this section, we provide different methods for ensuring responsible ML and we start by enlisting general responsible AI practices. 

\subsubsection{General Responsible AI Practices}
The following are some recommended AI practices to ensure effective and reliable AI systems\footnote{\url{https://ai.google/responsibilities/responsible-ai-practices/}}.

\begin{itemize}
    \item \textit{Consider human-centered design approach}: To have a large impact on the system being developed, it is important to consider the characteristics of the users for true recommendations.  
    \item \textit{Evaluate training and monitoring using suitable metrics}: Instead of using multiple metrics for evaluation of model training, ensure that the metric is appropriate for the context and goals of the systems and consider users' feedback in terms of surveys. 
    \item \textit{Examine your raw data}: The biases and abnormalities in the datasets (e.g., missing values, class imbalance, and incorrect labels)  are directly reflected by the learned ML models. To ensure the efficacy of the learning process, careful examination of the raw dataset is necessary while respecting the privacy concerns. 
    \item \textit{Understand limitations of the model and dataset}: It is crucial to understand the capability and limitations of the ML model and dataset, e.g., a model trained for detecting correlations cannot be used for inferences. 
    \item \textit{Repetitive Testing}: Once developed, ML systems should be tested again and again to ensure that they are working as intended. Rigorous tests should be performed to understand how the individual components of the ML system interact with each other. Other similar tests include testing for input drifts, using gold standard datasets, incorporating a larger sample base, and using quality checking mechanisms. 
    \item \textit{Continuous Monitoring and Updating}: To ensure the efficient performance of the ML systems deployed in real-time settings, continued monitoring and updating are required to identify and fix various issues encountered in realistic settings.  
\end{itemize}

\subsubsection{Responsible ML for Healthcare}
ML/DL techniques have a great potential for clinical applications (e.g., radiologist-level pneumonia detection \cite{rajpurkar2017chexnet} and dermatologist-level classification of skin cancer \cite{esteva2017dermatologist}, etc.) but their limited adoption in actual clinical settings indicates that these methods are not yet optimal and not ready for clinical deployment. In a recent study \cite{wiens2019no}, Wiens et al. have provided a roadmap towards safe, meaningful, and responsible ML for healthcare and argued that ML deployment in any field should be carried out by an interdisciplinary team that may include different stakeholders from multi disciplines, i.e., knowledge experts, decision-makers, and users. Examples for an interdisciplinary team having different stakeholders in the healthcare ecosystem are presented in Table \ref{tab:inter}. In addition, the authors also identified critical steps to be followed/considered when designing, testing, and deploying ML solutions for healthcare applications that include: (1) choosing the right problems; (2) developing a useful solution; (3) considering ethical implications; (4) rigorously evaluating the model; (5) thoughtfully reporting results; (6) deploying responsibly; and (7) making it to market. 

\begin{table}[!ht]
\centering
\caption{Examples for interdisciplinary teams having different stakeholders from multiple domains. (Adopted from \cite{wiens2019no})}
\begin{tabular}{|l|p{50mm}|}
\hline
\multicolumn{1}{|c|}{\textbf{Stakeholder Category}} & \multicolumn{1}{c|}{\textbf{Examples}}                       \\ \hline
\multirow{4}{*}{Knowledge experts}                  & Clinical experts, e.g., radiologist and dermatologists, etc. \\ \cline{2-2} 
                                                    & Health information and technology experts                    \\ \cline{2-2} 
                                                    & ML researchers, e.g., ML engineers and data scientists, etc. \\ \cline{2-2} 
                                                    & Implementation experts                                       \\ \hline
\multirow{4}{*}{Decision-makers}                    & Institutional leadership                                     \\ \cline{2-2} 
                                                    & Hospital administrators                                      \\ \cline{2-2} 
                                                    & State and federal government                                 \\ \cline{2-2} 
                                                    & Regulatory agencies                                          \\ \hline
\multirow{5}{*}{Users}                              & Physicians                                                   \\ \cline{2-2} 
                                                    & Nurses                                                       \\ \cline{2-2} 
                                                    & Laboratory technicians                                       \\ \cline{2-2} 
                                                    & Patients                                                     \\ \cline{2-2} 
                                                    & Care takers, e.g., friends and family                        \\ \hline
\end{tabular}
\label{tab:inter}
\end{table}

\subsection{Tools and Libraries for Secure and Private ML}
The main strength of ensuring secure ML relies on the development of security tools and algorithms. To ensure the security and privacy of ML models and data, various tools and libraries have been released so far. For example, \textit{TensorFlow Federated}\footnote{https://www.tensorflow.org/federated}, which is an open-source framework for distributed ML/DL that enables training of a global shared model in a federated environment without sharing clients' local data. CrypTen\footnote{https://github.com/facebookresearch/CrypTen} is a framework for secure and privacy-preserving ML built on PyTorch that provides secure computing techniques for ML/DL model training and inference using encrypted data and PyTorch-DP\footnote{https://github.com/facebookresearch/pytorch-dp}--a framework of PyTorch for training DL models with differential privacy. Similarly, \textit{OpenMined}\footnote{https://www.openmined.org/}-- an open-source community offers various tools and libraries for building privacy-preserving ML models which are briefly described below. 

\begin{itemize}
    \item \textit{PySyft}\footnote{https://github.com/OpenMined/PySyft} is python library for encrypted and privacy preserving ML. It extends PyTorch, TensorFlow, and Keras and supports differential privacy, federated learning, multi-party computation, and homomorphic encryption. 
    \item \textit{PyGrid}\footnote{https://github.com/OpenMined/PyGrid} is a platform built on PySyft that provides a peer-to-peer network to collectively train ML models.
    \item \textit{SyferText}\footnote{https://github.com/OpenMined/SyferText} is a privacy preserving framework for NLP tasks. 
\end{itemize}

\section{Open Research Issues}
\label{sec:open}
In this section, various open research issues related to the domain of secure, robust, and private ML for healthcare that require further research attention are presented.  

  
\subsection{Interpretable ML}
Although the advancement in ML/DL research has provided significant performance improvements over the previous state of the art methods in terms of performance metrics such as accuracy, precision, recall, and f1-measure, these advancements have made the learning process of modern models very complex and are usually deployed as a black-box. These black-box methods fail at providing rational or insights as well as at explaining their learning behavior and thought process for making predictions \cite{latif2019caveat}. The aforementioned problem is termed as the interpretability problem of ML in the literature, which is defined as the ability to describe the internal processes of an ML system in a human-understandable manner.

Moreover, interpretability of ML/DL techniques is required to ensure algorithmic fairness, robustness, and generalization based on potentially dispersed data collected from a heterogeneous population. This can eventually help in the smooth deployment and functionality of ML/DL systems in realistic settings. For a critical application like healthcare, the ML/DL model is expected to be highly accurate and understandable at the same time. Moreover, it has been argued that clinical integration of AI models will require interpretability \cite{jia2019clinical}. To perform an interpretation of ML models, questions about the fairness of model's predictions, transparency, and accountability are considered and interpretation is performed using explanation methods for justifying predictions of the model using visual, textual, or features information. Various methods have been proposed in the literature for explaining ML/DL models for general applications \cite{bach2015pixel,ribeiro2016should,lahav2018interpretable}, however, more research that is specifically focused on the interpretation of ML/DL systems used in healthcare applications is required.    


\subsection{Machine Learning on the Edge}
The advancements in ML research have revolutionized traditional healthcare (as discussed in earlier sections). Healthcare services will increasingly adopt the utilization of IoT devices and wearable sensors in the future, particularly with the evolution of smart cities and portable medical devices, e.g., portable MRI scanner. With such proliferation, there is a pressing need for pushing ML models training and inference on edge devices. This introduces unique challenges such as limited hardware and processing capabilities, etc. Moreover, this is crucial for portal medical devices that are utilized for patients in critical care as they cannot be moved to fixed medical equipment in the hospital. The research on enabling ML on edge devices (a.k.a fog) is in the early stages of development and requires further attention from the research community. The development of this field will enable to monitor patients in a critical situation and eventually enable continuous behavioral monitoring for improving individuals' life-style and timely detection of diseases. 


\subsection{Handling Dataset Annotation}
To increase the performance of ML/DL models, one natural strategy is to acquire more labeled training data. This requires that radiologists and medical experts spend their valuable time manually annotating medical data, e.g., medical images, signals, and reports. Another important aspect is devising true validation sets that will evaluate the performance of the ML/DL models and expose the limitations of these models. Therefore, manual annotation of samples into respective categories is time consuming, costly, and a tidy process. Automatic approaches should be developed to address this issue and one such technique is active learning which can be used to annotate unlabelled data samples. 

Data from multiple sources should be considered when performing annotation for specific clinical applications because single-source data might lack precise structured labels \cite{ghassemi2018opportunities}. The integration of multiple source data is an important application of ML in healthcare \cite{halpern2016electronic}, which is known as phenotyping \cite{richesson2013electronic}. NLP techniques and recurrent deep models can be used for extracting and integrating rich information from unstructured clinical notes to augment the capacity of data annotators. 

\subsection{Distributed Data Management and ML}
In healthcare settings, the data is generated in a distributed fashion, i.e., across different departments within a hospital and even across different hospitals. This necessitates the efficient management and sharing of distributed data for clinical analysis purposes, particularly using ML/DL models. In general, for developing ML/DL models, it is assumed that complete training and validation datasets are centrally available and easily accessible. Therefore, there is an increasing demand to develop methods for distributed data management and ML.  

\subsection{Fair and Accountable ML}
The literature on analyzing the security and robustness of ML/DL approaches reveals that the outcomes of these models lack fairness and accountability \cite{qayyum2019securing}. Whereas ensuring the fairness and accountability of predictions in life-critical applications like healthcare are of paramount importance, the \textit{fairness} property ensures that the ML model should not favor certain cases over others. Such discrimination mainly arises due to biases in the training data. On the other hand, \textit{accountability} property is concerned with the interpretation of the predictions. Fairness and accountability will assist in developing models robust to biases and imperfections such as past clinical practices 

\subsection{Model-Driven ML}

Although ML, AI, and big data are immensely useful tools for healthcare, these tools are not panacea and it is important to be aware of the associated caveats and pitfalls \cite{latif2019caveat}. Failing to realize this, one can easily fall prey to the dangerous dogma that data once available in abundance must and will speak for itself and can handle hypothesis generation as well---which in clinical terms would mean that data mining is sufficient and independent of the need of clinical interpretation, external validation, and understanding of data's provenance \cite{belgrave2017disaggregating}. To avoid the various problems that can arise from improper use of ML in healthcare, it is important to combine data-driven methods with hypothesis-driven or model-based methods (based on subject matter knowledge) and to bring scientific rigor in these studies. Properly designed experiments are also necessary for deriving causal explanations. Avenues for developing secure and robust ML solutions for healthcare that are scientifically robust and rigorous requires further attention from the community.

\section{Conclusions}
\label{sec:con}
The use of machine learning (ML)/deep learning (DL) models for clinical applications has great potential to transform traditional healthcare service delivery. However, to ensure a secure and robust application of these models in clinical settings, different privacy and security challenges should be addressed. In this paper, we provided an overview of such challenges by formulating the ML pipeline in healthcare and by identifying different sources of vulnerabilities in it. We also discussed potential solutions to provide secure and privacy-preserving ML for security-critical applications like healthcare. Finally, we presented different open research problems that require further investigation. 

\bibliographystyle{IEEEtran}

\end{document}